\documentclass[lettersize,journal]{IEEEtran}
\usepackage{amsmath,amsfonts}
\usepackage{algorithmic}
\usepackage{algorithm}
\usepackage{array}
\usepackage[caption=false,font=normalsize,labelfont=sf,textfont=sf]{subfig}
\usepackage{textcomp}
\usepackage{stfloats}
\usepackage{url}
\usepackage{verbatim}
\usepackage{graphicx}
\usepackage{cite}
\usepackage{multirow}
\usepackage{tabularx, makecell, booktabs}

\begin{document}

\title{Towards Camera Open-set 3D Object Detection
 for Autonomous Driving Scenarios}

\author{Zhuolin He, Xinrun Li, Jiacheng Tang, Shoumeng Qiu, Wenfu Wang, Xiangyang Xue, Jian Pu\textsuperscript{*}
\thanks{Manuscript received April 8, 2025. This work was supported in part by NSFC Projects (U62076067), Science and Technology Commission of Shanghai Municipality Projects (19511120700, 19ZR1471800), Shanghai Research and Innovation Functional Program (17DZ2260900), Shang-hai Municipal Science and Technology Major Project (2018SHZDZX01) and ZILab.} 
\thanks{Jian Pu is the corresponding author.}}

\markboth{Journal of \LaTeX\ Class Files,~Vol.~14, No.~8, August~2021}
{Shell \MakeLowercase{\textit{et al.}}: A Sample Article Using IEEEtran.cls for IEEE Journals}

\maketitle

\begin{abstract}
Conventional camera-based 3D object detectors in autonomous driving are limited to recognizing a predefined set of objects, which poses a safety risk when encountering novel or unseen objects in real-world scenarios. To address this limitation, we present OS-Det3D, a two-stage training framework designed for camera-based open-set 3D object detection. In the first stage, our proposed 3D object discovery network (ODN3D) uses geometric cues from LiDAR point clouds to generate class-agnostic 3D object proposals, each of which are assigned a 3D objectness score. This approach allows the network to discover objects beyond known categories, allowing for the detection of unfamiliar objects. However, due to the absence of class constraints, ODN3D-generated proposals may include noisy data, particularly in cluttered or dynamic scenes. To mitigate this issue, we introduce a joint selection (JS) module in the second stage. The JS module uses both camera bird's eye view (BEV) feature responses and 3D objectness scores to filter out low-quality proposals, yielding high-quality pseudo ground truth for unknown objects. OS-Det3D significantly enhances the ability of camera 3D detectors to discover and identify unknown objects while also improving the performance on known objects, as demonstrated through extensive experiments on the nuScenes and KITTI datasets.
\end{abstract}


\section{Introduction}
\IEEEPARstart{C}{onventional} camera 3D object detectors in autonomous driving are trained to recognize and locate a predefined set of objects \cite{li2022bevformer, li2023bevdepth}. These detectors operate under the assumption that they are in closed-world scenarios in which all possible object categories are known and labeled during training. However, the dynamic and unpredictable nature of real-world driving environments frequently introduces novel or unseen object categories, rendering these closed-set detectors inadequate \cite{cen2021open}. The inability to detect and respond to unknown objects poses a significant safety risk \cite{bozhinoski2019safety}, and autonomous driving systems still frequently demonstrate erroneous or unexpected behaviors that could lead to undesirable outcomes \cite{9284628}, thereby increasing the need for camera open-set 3D object detection.

\begin{figure}[t]
\centering
  \includegraphics[width=0.9\columnwidth]{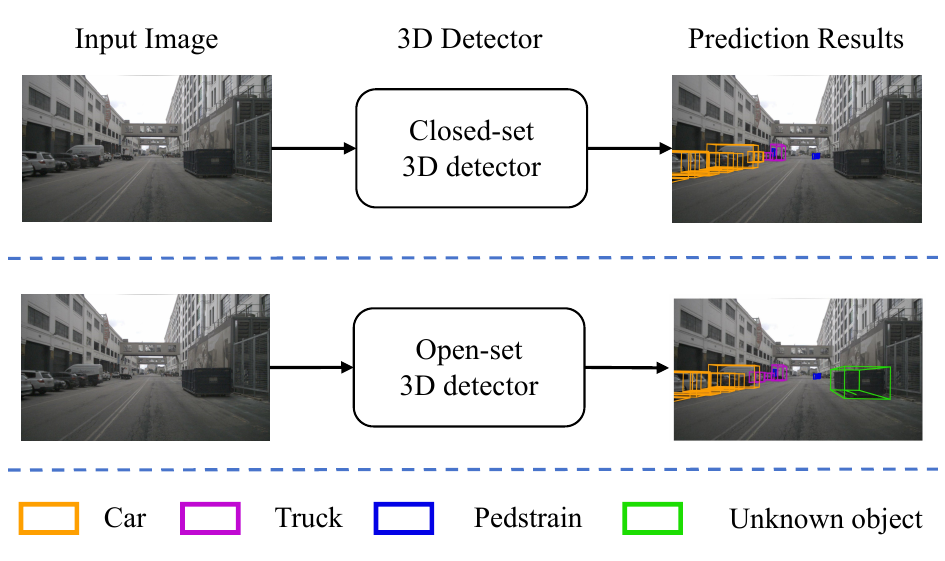}
  \caption{ \textbf{ Prediction with open-set 3D object detection.} While closed-set 3D detection fails to detect unknown objects on the road, open-set 3D detection is able to identify both known and unknown objects.}
  \label{fig:exp}
\end{figure}

Open-set scenarios introduce new challenges to the 3D object detection pipeline. The existing methods \cite{gupta2022ow, joseph2021towards} address similar challenges in the 2D domain and typically follow a two-stage paradigm: first, object proposals are generated, and then, unknown instances are identified from the proposals. However, extending this paradigm from 2D to 3D introduces significant additional challenges. Generating accurate 3D object proposals from images is inherently difficult because RGB-based models tend to overfit to shortcut cues such as textures and discriminative parts \cite{huang2023good} while lacking reliable depth information as compared with LiDAR.
Furthermore, although the objectness scores provided by the proposed methods are effective in terms of indicating the presence of objects, high-confidence proposals tend to correspond to known instances seen during training, and directly using such noisy pseudo-labels for training can limit overall performance \cite{wang2021combating}.

To overcome these limitations, we propose OS-Det3D, a training framework that combines complementary information from camera and LiDAR data to enable open-set 3D object detection in two stages.
The first stage, driven by the 3D object discovery network (ODN3D), focuses on generating general 3D object proposals by leveraging geometric cues from LiDAR point clouds.
However, simply relying on existing learning-based proposal methods \cite{yin2021center,zhang2023simple} is not well suited for open-set scenarios, as these methods typically formulate the task as a binary foreground/background classification problem. As a result, the models overfit to labeled objects and treat unlabeled objects as background, which hinders their generalization to unseen objects \cite{kim2022learning}.
To address these issues, ODN3D introduces a novel geometric-only Hungarian (GeoHungarian) matching algorithm to mitigate overfitting to labeled instances and combines it with a 3D objectness score to guide the network toward learning class-agnostic geometric features that generalize well to unseen objects.

The second stage focuses on identifying unknown instances from the generated proposals. While GT filtering is effective in 2D open-set methods \cite{joseph2021towards, gupta2022ow} in terms of removing known objects from the proposal pool, it becomes less reliable in the 3D domain, and in real-world autonomous driving scenarios, ground truth annotations often fail to cover all known objects. Training with such noisy pseudo-labels could negatively impact the detection performance for both known and unknown objects. To address this issue,
we propose a novel joint selection module that computes a hybrid score by combining the 3D objectness scores from ODN3D with the 
BEV feature responses from the camera detector, enabling the selection of objects with novel appearances. The 3D objectness scores help indicate the likelihood that a proposal corresponds to an object, while the BEV feature responses reflect the appearance similarity to known categories. This integration significantly enhances the ability to filter noisy proposals and accurately identify potential unknown objects, and these selected proposals then serve as pseudo ground truth for training the camera 3D detector, enhancing its ability to detect unknown objects.

Through extensive experiments on the nuScenes \cite{nuscenes2019} and KITTI \cite{geiger2012we} datasets, we demonstrate the efficacy of OS-Det3D in enabling camera 3D detectors to successfully identify unknown objects while maintaining or even improving their performance on known objects. Notably, our proposed ODN3D achieves significant improvements in terms of the Recall and AP values for unknown object discovery, outperforming the state-of-the-art open-set 3D detection methods on the KITTI dataset.

Our contributions can be summarized as follows:

\begin{itemize}
    \item We introduce ODN3D, a novel 3D object proposal network that effectively learns geometric cues for discovering novel 3D objects by combining GeoHungarian matching and the 3D objectness score.

\item We propose a joint selection module that leverages cross-modality information to refine the selection of the pseudo ground truth for unknown objects, ensuring effective learning and generalization.

\item We present OS-Det3D, a novel two-stage training framework that enables cameras to perform open-set 3D object detection, marking a significant advancement in the field of autonomous driving perception.
\end{itemize}

\section{Related Work}

\subsubsection{Closed-set Camera 3D Object Detection}
Closed-set 3D object detection via cameras has undergone rapid advancements in recent years. Unlike LiDAR-based approaches \cite{zhou2018voxelnet,shi2020pv,deng2021voxel}, which provide precise depth information through direct distance measurement, camera-based 3D object detection \cite{10274675, 10061347} relies solely on visual cues such as texture, color, and scene structure to estimate object depth and 3D position. One major line of research in camera-based 3D detection focuses on utilizing bird's eye view (BEV) representations. Methods such as BEVFormer \cite{li2022bevformer} and BEVDepth \cite{li2023bevdepth} have shown that transforming image features into a BEV space allows for more accurate spatial reasoning, which is crucial for 3D object localization. However, despite these advancements, camera-based methods remain constrained by their closed-set nature and are trained to detect only predefined objects. This restriction poses a significant challenge in open-set scenarios, where previously unseen objects may emerge, leading to difficulties in accurate identification and localization.

\begin{figure*}[t]
\centering
  \includegraphics[width=0.9\textwidth]{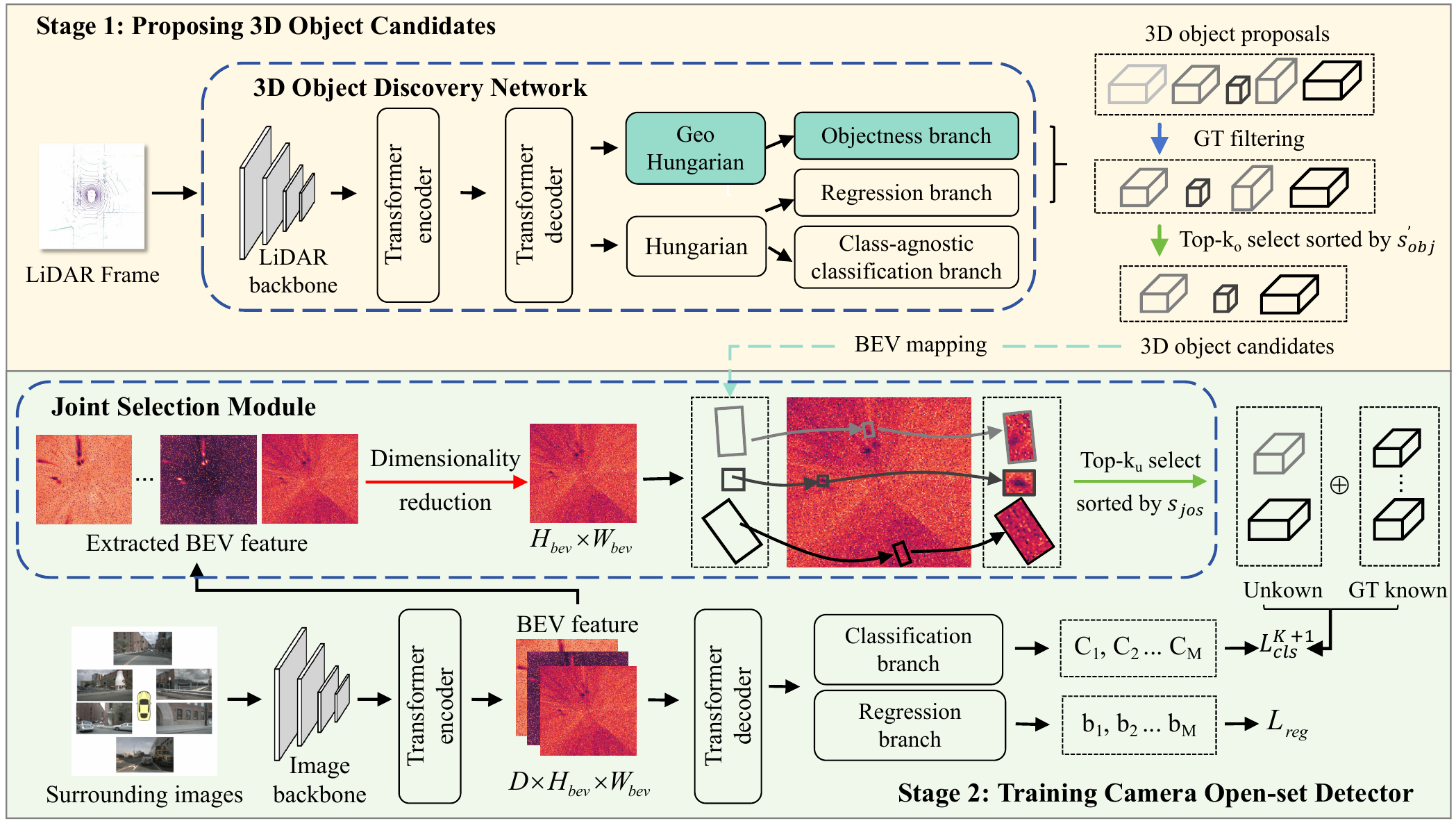}
  \caption{ \textbf{Proposed OS-Det3D training framework.} For stage 1, we have designed a 3D object discovery network to propose 3D object candidates via an encoder-decoder architecture. At the decoder output, each 3D object prediction process involves two matching strategies. The objectness branch estimates the localization quality via a 3D objectness score, $s^{\prime}_{obj}$, where the output of ODN3D is a set of 3D object proposals, each with a predicted $s^{\prime}_{obj}$. In the ground truth (GT) filtering step, we remove 3D object proposals that overlap with known category instances and select the top-$k_{o}$ proposals as 3D object candidates. In stage 2, we extract bird's eye view (BEV) features from the camera encoder output, perform dimensionality reduction, and compute the BEV feature responses for each 3D candidate. By combining the BEV feature responses with the 3D objectness score, we define a joint selection score, $s_{jos}$, to rank and select the top-$k_{u}$ candidates as the pseudo-GT for unknown objects. These pseudo-unknowns are then concatenated with the known ground truth and input into the camera detector's classification branch for training.
  }
  \label{fig:pipeline}
\end{figure*}

\subsubsection{Class-agnostic Region Proposal Network} Proposal-based methods can be categorized into two types: learning-free \cite{uijlings2013selective, 7938411, wang20224d} and learning-based approaches \cite{yin2021center,zhang2023simple}. Learning-free methods rely on fixed geometric rules and hand-crafted features, which can result in missed detections or false positives in complex scenarios \cite{o2015learning}, making them ill-suited for open-set detection. As a result, we adopt a learning-based approach that enables the model to learn generalized object representations and produce flexible and reliable responses to previously unseen objects.
Early works in 2D \cite{alexe2012measuring, zitnick2014edge} emphasized category-independent proposals, aiming to identify all objects in an image regardless of their category. The core issue in 2D object detection is that models often treat unannotated objects as background, suppressing their detection during inference \cite{wang2019region}. To solve this issue, the OLN \cite{kim2022learning} introduced a geometric-only approach that focuses on learning objectness cues, thereby facilitating the detection of unannotated novel objects. GOOD \cite{huang2023good}, paralleling the OLN, advanced the geometric-only approach by incorporating geometric information such as depth for enhanced generalization of the detector. While these advancements in 2D are significant, applying geometric-only approaches in 3D environments presents limitations.
To discover novel unknown objects with 3D regions, recent works \cite{lu2023open, cao2024coda} have employed a LiDAR 3D object detector as a class-agnostic 3D detector with binary classification. While class-agnostic 3D detectors efficiently detect novel objects indoors, they often miss novel unknown objects in autonomous driving scenarios. This failure occurs because class-agnostic 3D detectors rely on learning classifications from training categories, making them prone to overfitting to the labeled known objects and limiting their ability to detect novel unknown objects. In this work, we propose incorporating geometric cues by measuring the 3D localization quality in a class-agnostic 3D detector, demonstrating an improved performance in terms of discovering novel unknown objects.

\subsubsection{Open-set and Open-World Object Detection}
The concept of the open-set object detection problem was first introduced in early work \cite{dhamija2020overlooked} and has been further explored in subsequent research \cite{zheng2022towards, sarkar2024open}. In open-set settings, the model is provided with labels for certain known category instances during training but is required to detect both known and unknown objects during testing
. Early works, such as MC-dropout \cite{miller2018dropout} and Bayesian methods \cite{pham2018bayesian}, directly distinguished unknown objects from closed-set detection results. Extending beyond open-set object detection, the ORE \cite{joseph2021towards} introduced a novel setting called open-world object detection that aims to enable models to recognize objects of known categories while detecting and incrementally learning to identify previously unseen new categories.
To further advance open-world object detection, the OW-DETR \cite{gupta2022ow} introduced the open-world detection transformer, which uses an attention-driven pseudo-labeling scheme to supervise unknown object detection. In this scheme, unmatched object proposals with high backbone activation are selected as unknown objects. Furthermore, by shifting the focus to 3D object detection, the OSIS \cite{wong2020identifying} first introduced the open-set 3D instance segmentation network with LiDAR, which is uniquely designed for 3D space open-set conditions, and the MLUC \cite{cen2021open} focuses on identifying unknown objects from closed-set detection predictions by employing a LiDAR detector. However, beyond these studies, exploration into camera open-set 3D detection remains significantly underdeveloped.

\section{Proposed Method}

Fig.~\ref{fig:pipeline} illustrates the overall architecture of our OS-Det3D training framework. We empower a closed-set camera 3D detector with open-set detection capabilities through two innovative components: a 3D object discovery network (ODN3D)
 that identifies general 3D objects and a joint selection module that distinguishes unknown 3D objects from these initial detections. The entire framework undergoes a two-stage training process for optimal performance.

 \subsection{Preliminaries}
\label{sec:PF}
 Let $\mathcal{K} = \{1, 2, \ldots, C_{K}\}$ represent the set of known object categories, where $C_{K}$ is the number of known categories. $\mathcal{U} = \{C_{K}+1, \ldots, C_{K}+C_{U}\}$ denotes the set of unknown classes, where $C_{U}$ is the number of unknown categories. In an open-set condition, the training dataset $\mathcal{D}_{\text{train}}$ consists only of known categories $\mathcal{K}$, whereas the test dataset $\mathcal{D}_{\text{test}} = \mathcal{K} \cup \mathcal{U}$ includes both known and unknown categories. The test dataset classes are denoted by $\{1, 2, \ldots, C_{K}, C_{K}+1, C_{K}+2, \ldots, C_{K}+C_{U}\}$.
The training dataset $\mathcal{D}_{\text{train}} = \{\mathcal{I}, \mathcal{Y}\}$ contains images $\mathcal{I} = \{I_{1}, \ldots, I_{m}\}$ with 3D annotations $\mathcal{Y}_{\text{train}} = \{\boldsymbol{b}_{i} \mid i = 1, 2, \ldots, k\}$, where each annotation $\boldsymbol{b}_{i} = \left[c, x, y, z, w, l, h, r\right]$. In these annotations, $c$ represents the known class in $\mathcal{D}_{\text{train}}$, the 3D bounding box center coordinates $(x, y, z)$, the dimensions $(w, l, h)$, and the rotation (yaw angle) $r$.
During training and testing, all categories in $\mathcal{U}$ are collectively treated as a single category. Furthermore, instances from $\mathcal{U}$ are not annotated during training, and the model is trained exclusively using instances from $\mathcal{K}$ \cite{cen2021open}.
Additionally, a camera open-set 3D detector is trained to detect unknown objects from unknown categories $\mathcal{U}$ while recognizing known objects in $\mathcal{K}$.

\subsection{3D Object Discovery}
 \label{sec:rpn}

The key challenge in proposing 3D object candidates in autonomous driving scenarios is effectively generalizing them to unknown objects from a limited set of known classes. Recent works \cite{lu2023open,cao2024coda} have used a class-agnostic 3D detector that replaces multiclass classification with binary classification to propose 3D object candidates indoors. However, this type of class-agnostic 3D detector struggles with generalization in autonomous driving scenarios and is prone to overfitting to labeled categories. While 2D geometry-only methods \cite{kim2022learning, huang2023good} have demonstrated improved generalizability by learning geometric cues, they unfortunately fail to be applicable in 3D environments.

To propose 3D candidates for novel unknown objects, we propose ODN3D, which incorporates a geometry-only branch with a newly designed GeoHungarian matching and a 3D objectness score. GeoHungarian matching is a geometry-only bipartite matching strategy that focuses on localization cues (location, scale) independent of classification, enhancing the model's ability to generalize to unlabeled objects, and the 3D objectness score aims to measure the localization quality of 3D candidates by appropriately describing the location and scale measurements. The ODN3D network is shown in stage 1 of Fig. \ref{fig:pipeline}.

\subsubsection{GeoHungarian Matching} While Hungarian matching \cite{wang2022detr3d} establishes a correspondence between the ground truth and predictions, its inclusion of classification costs tends to penalize the detection of unlabeled unknown objects. To eliminate dependence on labels, we introduce GeoHungarian matching, which focuses solely on geometric information. The geometry-only bipartite matching problem is reformulated as follows:
\begin{equation}
  \sigma^{\prime} = \mathop{\arg\min}\limits_{\sigma \in \mathcal{P}} \sum_{j=1}^{M} \mathcal{L}_{box}(b_j,\hat{b}_{\sigma(j)}),
  \label{eq:important}
\end{equation}
where $\mathcal{P}$ denotes the set of permutations, $M$ denotes the number of ground truth boxes, $b$ denotes the geometric annotation of $\left[x, y, z, w, l, h, r\right]$, $\sigma(*)$ returns the corresponding index of the ground-truth bounding box, and $\mathcal{L}_{box}$ is the \emph{L}1 loss for the bounding box parameters. The 3D object predictions that closely match the ground truth in geometric terms are presented as positive samples.

\begin{figure}[t]
\centering
  \includegraphics[width=0.8\columnwidth]{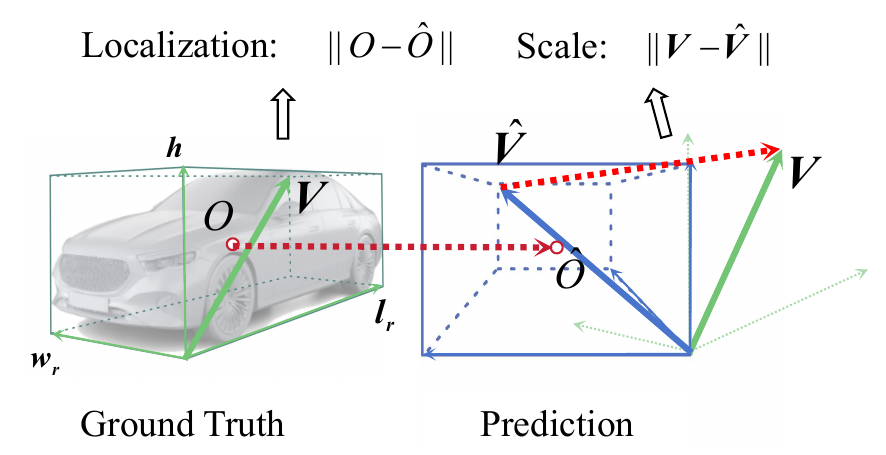}
  \caption{ \textbf{3D localization quality measurement.} The geometric features of a GT 3D box are divided into two parts: location $(x,y,z)$ and scale $(w,h,l,r)$. These parts can be represented as center points $O$ for localization and vectors $v$ for scale, while the corresponding predictions are $\hat{O}$ and $\hat{\boldsymbol{v}}$. Considering the inconsistency in scale, we reformulated the vector $\boldsymbol{v}$ into a matrix $\boldsymbol{V}$, providing a more accurate description of the 3D box's scale information. We then calculate the corresponding predicted scale distance.
  }
  \label{fig:3ds}
\end{figure}

\subsubsection{3D Objectness Score}
 Incorporating localization quality is not entirely new in 2D object detection; the OLN \cite{kim2022learning} adopts the centerness \cite{tian2019fcos} and IoU \cite{jiang2018acquisition} scores for location and shape quality measures, respectively, whereas the objectness score of the OLN is computed as a geometric mean of the centerness and IoU scores.

However, the IoU score has limitations in terms of capturing subtle geometric variations, as small deviations in shape or orientation do not significantly alter the overlap volume. This phenomenon can result in reduced accuracy and stability, as shown in Table \ref{tab:object-score}. In our experiments, we found that more precisely and consistently defined geometric cues lead to better generalization of object proposals. Inspired by this observation, we designed a 3D objectness score adapted for 3D environments.

\begin{figure}[t]
  \centering
  \includegraphics[height=4.2cm]{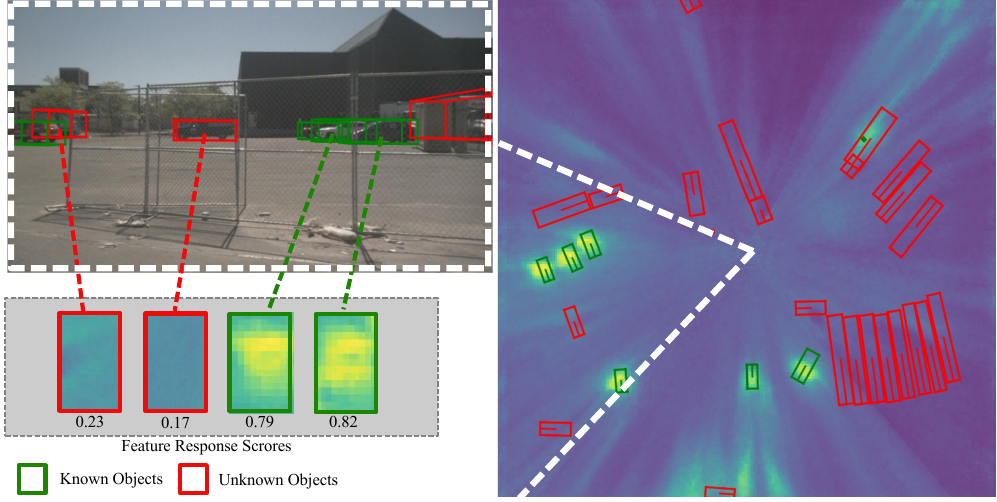}
  \caption{Visual illustration of PCA-transformed BEV feature activation (sample: 52d4b1f9e13a46f69257df0b1d68e1ec). The images on the right depict PCA-transformed BEV feature activation extracted from BEVFormer and trained on nuScenes Split 2. The white area on the top left corresponds to the white area in the right image. The red boxes correspond to the unknown GT, and the green boxes correspond to the known GT. The BEV feature response scores for the boxes can be calculated from the mean responses within the corresponding BEV window.}
  \label{fig:att_fea_compute}
\end{figure}

\label{JOS}
\begin{figure*}[ht]
  \centering
  \includegraphics[width=0.9\textwidth]{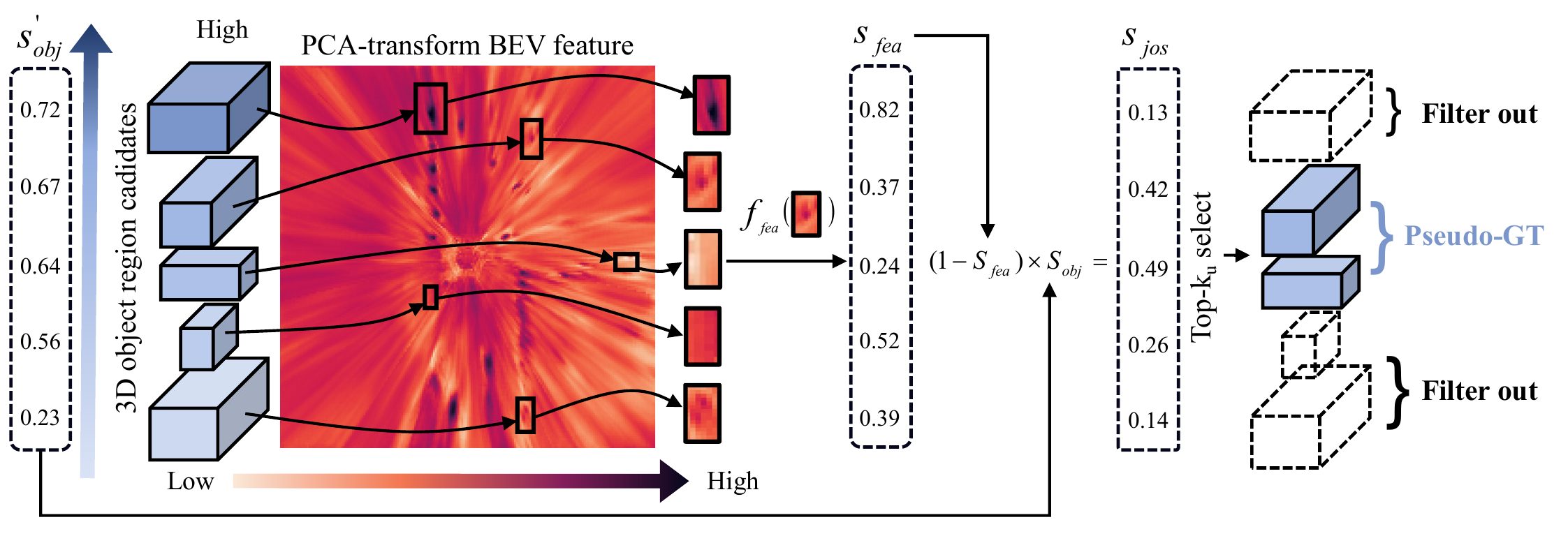}
  \caption{\textbf{Overview of the joint selection module}. The joint selection module computes the corresponding feature response score $s_{fea}$ for the $\text{top-}k_{o}$ candidates as the mean feature score within a BEV 2D region of interest. Then, we select the $\text{top-}k_{u}$ candidates sorted by $s'_{obj}$ as pseudo unknown ground truth.
  }
  \label{fig:JOS}
\end{figure*}

A 3D bounding box $\boldsymbol{b}_{i} = \left[c, x, y, z, w, l, h, r\right]$ can be divided into a localization part $(x, y, z)$ and a scale part $(w, l, h, r)$, disregarding class $c$. Considering that localization differences between the ground truth and predicted boxes are greater than scale differences and that each has distinct physical meanings, we design a 3D objectness score that encompasses measurements of both 3D localization and scale, as shown in Fig. \ref{fig:3ds}.

For the localization measurement, we calculate the \emph{L}1 distance between the center point $O = (x,y,z)$ of the ground truth 3D bounding box and the center point $\hat{O}$ of the 3D object predictions.
For scale measurement, to address the inconsistency in units between $r$ (measured in radians) and $w, l, h$ (measured in meters), we reformulate the scale information as $ \boldsymbol{l} = \left[l,0,0\right]^\top$, $\boldsymbol{w} = \left[0,w,0\right]^\top$, and $\boldsymbol{h} =\left[0,0,h\right]^\top$. The yaw angle $r$ can be represented by the rotation matrix $\boldsymbol{R_z}$. By applying $\boldsymbol{R_z}$ to $\boldsymbol{l}$ and $\boldsymbol{w}$, we obtain the rotated vectors $\boldsymbol{l_{r}} = \boldsymbol{R_z} \cdot \boldsymbol{l}$ and $\boldsymbol{w_{r}} = \boldsymbol{R_z} \cdot \boldsymbol{w}$, effectively aligning them with the 3D object's orientation. The scale of the ground truth 3D bounding box can be described as a matrix $\boldsymbol{V}=\left[\boldsymbol{l_{r}},\boldsymbol{w_{r}},\boldsymbol{h} \right]^\top$, and the prediction 3D bounding box is denoted by $\hat{\boldsymbol{V}}$. We use the \emph{L}1 distance between matrices $\boldsymbol{V}$ and $\hat{\boldsymbol{V}}$ as a measure of scale. The centerness score and scale score of the 3D prediction are calculated as follows:
\begin{equation} 
s_{center} =  \phi_1 \left(\vert\vert O-\hat{O} \vert\vert\right),
\end{equation} 
\begin{equation}
s_{scale} = \phi_2 \left(\vert\vert\boldsymbol{V}-\hat{\boldsymbol{V}}\vert\vert\right).
\end{equation} 
Here, $\phi_1$ and $\phi_2$ are two Gaussian kernel functions that can be written as
$
    \exp\left( { - {\vert\vert \textbf{x}-\hat{\textbf{x}}\vert\vert^2} / { 2 \tau }}\right),
$
where $\textbf{x}$ and $\hat{\textbf{x}}$ are two distribution vectors, and the parameter $\tau$ is used to normalize the centerness score and scale score.

With the ground truth objectness scores set to 1, we can represent $s_{obj}=\sqrt{s_{center} \cdot s_{scale}}$, which yields a score ranging from 0 to 1. This score reflects the quality of the geometric evaluation, with values closer to 1 indicating better quality. The objectness loss of our objective branch in the form of \emph{L}1 can be written as follows:
\begin{equation} 
\mathcal{L}_{obj} = \sum_{j=1}^{M} \mathcal{L}_{L1}\left(s_{obj(j)},\hat{s}_{obj(\sigma^{\prime}(j))}\right).
\end{equation} 

\subsubsection{Hungarian Matching} In 3D space, predictions are much sparser than they are in 2D space, making it challenging for an initial imprecise prediction to be successfully aligned with the ground truth during training when relying solely on a geometry-based matching strategy. This misalignment can hinder model convergence. To address this issue, we retain the Hungarian matching algorithm and its corresponding branches in ODN3D during the training phase, as illustrated in Fig. \ref{fig:pipeline}. During inference, the class-agnostic classification branch is omitted. Our experiment in Tab. \ref{tab:LiDAR-rpn-ab} shows that maintaining Hungarian matching and its corresponding branches ensures stable model performance.

\subsection{Joint Selection Module}
\label{sec:JOS}

The quality of pseudo-labels plays a critical role in the detector's ability to recognize unknown objects. While prior 2D open-set methods \cite{vaze2022openset, wu2022uc} have focused primarily on distinguishing unknown instances based on visual appearance or feature-level cues, our goal is to select high-quality pseudo-labels from generic 3D proposals that not only exhibit a novel appearance but also possess accurate 3D localization (e.g., position, size, and orientation). Poor 3D localization directly degrades the detection performance, whereas excessive similarity to known appearances reduces the model's generalizability to novel categories.

Higher objectness scores indicate that a candidate has a better location quality\cite{kim2022learning}, and the magnitude of image feature activation helps indicate the presence of known or similar objects \cite{yu2024emergence, caron2021emerging}.
Therefore, 3D object region candidates with high objectness scores and low BEV feature responses are more likely to meet our criteria for selection as potential unknown objects, as shown in Fig. \ref{fig:att_fea_compute}. We can compute the BEV feature score for a region proposal by calculating the responses within its corresponding BEV window, and the joint selection score can be represented as follows:
\begin{equation}
    s_{jos}=s^{\prime}_{obj} \cdot (1 - s_{fea}),
\end{equation}
where
\begin{align}
    s_{fea} &= \frac{1}{\text{w} \times \text{l}} \sum_{i=1}^{\text{w}} \sum_{j=1}^{\text{l}} \boldsymbol{A} \left(x + i \cdot \cos(r)        - j \cdot \sin(r), \right. \notag \\
            &  \quad \left. y + i \cdot \cos(r) + j \cdot \sin(r) \right).
  \label{eq:att_score}
\end{align}
$\text{w}=int(w)$ and $\text{l}=int(l)$, where $w$ and $l$ represent the width and length of the 3D box, respectively. The corresponding BEV feature $\boldsymbol{A} \in \mathbb{R}^{H_{bev} \times W_{bev}}$ refers to the dimensionally reduced version of the original BEV features $\boldsymbol{A} \in \mathbb{R}^{C \times H_{bev} \times W_{bev}}$.
Using original BEV features can result in inconsistent feature responses across channels $C$, with response values for known or similar objects varying significantly between channels, making it difficult to obtain representative values. We use principal component analysis \cite{wold1987principal} for dimensionality reduction, or alternatively, we use mean pooling for dimensionality reduction and apply normalization, both of which yield similar results. This observation suggests that the magnitude of BEV feature responses indicates the presence of known or similar objects at specific spatial positions, which helps distinguish between known and unknown objects.
Finally, we rank these $k_{o}$ candidates based on $s_{jos}$ and select the $\text{top-}k_{u}$ candidates as pseudo-GT $\mathcal{Y}_{\text{pseudo-GT}}$.

We provide an illustrative example of the joint selection procedure, as depicted in Fig. \ref{fig:JOS}. The two figures differ in terms of their respective viewpoints. In this procedure, the joint selection module receives 3D object region candidates as inputs and computes the average attention score, denoted by $s_{fea}$, over the BEV 2D regions corresponding to these candidates.

\subsection{Two-stage training strategy}
\label{sec:two-stage}

Algorithm~\ref{alg:osdet3d} clearly summarizes the overall training procedure of our proposed two-stage framework. Note that during inference, BEVFormer utilizes only camera data.

\subsubsection{First Stage} In the first stage, BEVFormer and ODN3D are trained independently with instances of known classes. The loss formulation used in this stage is defined as follows:
\begin{equation}
  \mathcal{L}_{Camera} = \mathcal{L}^{\mathcal{K}}_{cls} + \mathcal{L}_{reg}, 
\end{equation}
\begin{equation}
  \mathcal{L}_{LiDAR} = \mathcal{L}^{\{0,1\}}_{cls}+\mathcal{L}_{reg}+\mathcal{L}_{obj},
  \label{eq:stage1_LiDAR_loss}
\end{equation}
where the loss terms for classification, bounding box regression and objectness scoring are denoted by $\mathcal{L}_{cls}$, $\mathcal{L}_{reg}$ and $\mathcal{L}_{obj}$, respectively. The standard focal loss \cite{lin2017focal} is employed when formulating the classification loss. The term $\mathcal{L}^{\mathcal{K}}_{cls}$ refers to $\mathcal{L}_{cls}$ when training to classify each known class $\mathcal{K}$, whereas $\mathcal{L}^{(\mathcal{K})}_{cls}$ signifies the binary classification loss $\mathcal{L}_{cls}$ in which known classes $\mathcal{K}$ are grouped into a single category. $\mathcal{L}_{reg}$ and $\mathcal{L}_{obj}$ are formulated by using the standard \emph{L1} regression loss.
ODN3D proposes a set of 3D object region proposals.
Each proposal includes a predicted objectness score $s^{\prime}_{obj}$.
After GT filtering, we select the $\text{top-}k_{o}$ proposals as 3D object region candidates, ranked by $s^{\prime}_{obj}$.

\subsubsection{Second Stage} The second stage involves only the training of BEVFormer. Before training, we reload the pretrained weights of BEVFormer in Stage 1, excluding the classification branch. As mentioned above, the joint selection module selects a set of pseudo-GT $\mathcal{Y}_{\text{pseudo-GT}}$.
The original training label set ($\mathcal{Y}_{\text{train}} = \mathcal{Y}_{\text{GT}}$) is augmented by combining it with the selected pseudo-GT set to form the updated training set $\mathcal{Y}_{\text{train}} = \mathcal{Y}_{\text{GT}} \cup \mathcal{Y}_{\text{pseudo-GT}}$.
In this stage, we train the camera 3D object detector by using the following loss formulation:
\begin{equation}
  \mathcal{L}_{camera} = \mathcal{L}^{\mathcal{K}+1}_{cls} + \mathcal{L}_{reg},
\end{equation}
where 
\begin{equation}
  \mathcal{L}^{\mathcal{K}+1}_{cls} = \mathcal{L}^{\mathcal{K}}_{cls} + \sum^{k_{u}}_{i=1}s^{\prime}_{{obj}_{(i)}} \mathcal{L}^{\mathcal{(U)}}_{cls}(c_i,\hat{c}_{\sigma^{*}(i)}).
  \label{eq:cam_loss}
\end{equation}
The term ${\mathcal{(U)}}$ refers to unknown classes $\mathcal{U}$ that are grouped into a single category, and 
the weight of the unknown classification loss is determined via the objectness score $s^{\prime}_{obj}$. The advantage of this process is that it allows the model to focus more on regions with higher objectness, thereby improving the detection accuracy of unknown objects while reducing the impact of noisy or less confident regions.

\begin{algorithm}[t]
\caption{OS-Det3D workflow.}
\label{alg:osdet3d}
\renewcommand{\algorithmicrequire}{\textbf{Input:}}
\renewcommand{\algorithmicensure}{\textbf{Output:}}

\begin{algorithmic}
\STATE \textbf{Stage 1: Class-Agnostic Proposal Generation}
\STATE \textbf{Input:} camera images $\mathcal{I}$, LiDAR point clouds $\mathcal{X}$, ground truth labels for known objects $\mathcal{Y}_{\text{GT}}$
\STATE \textbf{Output:} a closed-set camera 3D object detector $\mathcal{F}_{close}$, object proposal set $\mathcal{O}$
\STATE \quad $\mathcal{F}_{close} \leftarrow (\mathcal{I}, \mathcal{Y}_{\text{GT}})$
\STATE \quad ODN3D $ \leftarrow (\mathcal{X}, \mathcal{Y}_{\text{GT}})$
\STATE \quad $\mathcal{O}$ = ODN3D $(\mathcal{X}) $

\vspace{2mm}
\STATE \textbf{Stage 2: Training the Camera Open-set Detector}
\STATE \textbf{Input:} init $\mathcal{F}_{open}$ with $\mathcal{F}_{close}$, $\mathcal{I}$,  $\mathcal{Y}_{\text{GT}}$, $\mathcal{O}$
\STATE \textbf{Output:} an open-set camera 3D object detector $\mathcal{F}_{open}$
\FOR { $(img, proposals)$ in DataSample($\mathcal{I}$, $\mathcal{O}$)}
    \STATE $f _{\text{BEV}}$ = ExtractBEVFeature($img$)
    \STATE $\mathcal{Y}_{\text{pseudo-GT}}$ = JointSelection$(proposals, f_{\text{BEV}})$
    \STATE  $\mathcal{Y}_{\text{train}} = { \mathcal{Y}_{\text{pseudo-GT}} \cup \mathcal{Y}_{\text{GT}}}$
    \STATE $\mathcal{F}_{open} \leftarrow ( img, \mathcal{Y}_{\text{GT}} ) $
\ENDFOR

\vspace{2mm}
\STATE \textbf{Inference:} Results = $\mathcal{F}_{open}(\mathcal{I})$

\end{algorithmic}
\end{algorithm}

\section{Experiments}

\begin{table*}[t]
\centering
\caption{Dataset split composition in the evaluation. The semantics and the number of frames and instances (objects) across the splits of the two datasets are shown.}
{\footnotesize\setlength{\tabcolsep}{.8mm}
\begin{tabular}{l|l|cc|cc|cc}
\toprule
{\textbf{Dataset}} & {\textbf{Dataset Split}}          & \multicolumn{2}{c|}{{\textbf{KITTI Split}}}   & \multicolumn{2}{c|}{{\textbf{nuScenes Split 1}}}  & \multicolumn{2}{c}{{\textbf{nuScenes Split 2}}} \\ 
\midrule
              &     & \multicolumn{1}{c|}{Known}  & Unknown   & \multicolumn{1}{c|}{Known}  & Unknown & \multicolumn{1}{c|}{Known}  & Unknown  \\ \cline{3-8} 
& {Semantic split}     & \multicolumn{1}{c|}{\begin{tabular}[c]{@{}c@{}}Car, Bicycle, \\ Pedestrian\end{tabular}} & Van,Truck & \multicolumn{1}{c|}{\begin{tabular}[c]{@{}c@{}}Car, Bicycle, \\ Pedestrian\end{tabular}} & \begin{tabular}[c]{@{}c@{}}Barrier, Bus, Debris \\ Construction  vehicle, \\ Truck, Traffic cone, \\ Trailer, Motorcycle\end{tabular} & \multicolumn{1}{c|}{\begin{tabular}[c]{@{}c@{}}Car, Bicycle,\\Construction vehicle,\\ Pedestrian, Barrier, \end{tabular}} & \begin{tabular}[c]{@{}c@{}}Bus,Truck, Trailer, \\ Traffic cone, \\ Motorcycle, Debris\end{tabular} \\ 
\midrule
& {Scenes}    & \multicolumn{2}{c|}{-}  & \multicolumn{2}{c|}{{700}}  & \multicolumn{2}{c}{{700}}     \\
Training & {Frames}     & \multicolumn{2}{c|}{{7481}}   & \multicolumn{2}{c|}{{28130}}   & \multicolumn{2}{c}{{28130}}  \\ 
\cmidrule{2-8}
& {Instances} & \multicolumn{1}{c|}{{8690}}  & -         & \multicolumn{1}{c|}{{608643}}  & -        & \multicolumn{1}{c|}{{745731}}    & -      \\ 
\midrule
& {Scenes}        & \multicolumn{2}{c|}{-}  & \multicolumn{2}{c|}{{150}}     & \multicolumn{2}{c}{{150}}  \\
Testing & {Frames}         & \multicolumn{2}{c|}{{3769}}  & \multicolumn{2}{c|}{{6019}}  & \multicolumn{2}{c}{{6019}}  \\ 
\cmidrule{2-8}
& {Instances}     & \multicolumn{1}{c|}{{4845}}      & {1255}      & \multicolumn{1}{c|}{{116732}}     & {70796}    & \multicolumn{1}{c|}{{146102}}    & {41426}    \\ 
\bottomrule
\end{tabular}
}
\label{tab:task_split}
\end{table*}

\begin{table*}[t]
\centering
\caption[Valori medi]{Camera open-set 3D object detection results on the nuScenes validation dataset. The (supervised) method includes the unknown classes in the training set, serving as the upper bound for the open-set detection performance.}
{\footnotesize\setlength{\tabcolsep}{.8mm}
{\begin{tabular}{l|ccc|ccc}
\toprule[1.2pt]                      
\multirow{2}{*}{\textbf{Methods}} & \multicolumn{3}{c|}{\textbf{{nuScenes Split 1 test set}}}  & \multicolumn{3}{c}{\textbf{{nuScenes Split 2 test set}}}    
\\ 
          & $\text{AR}_{\texttt{unk}} \uparrow $      & $\text{AP}_{\texttt{unk}} \uparrow $          & ${\text{mAP}_{\texttt{known}}} \uparrow $         & $\text{AR}_{\texttt{unk}} \uparrow $         & $\text{AP}_{\texttt{unk}} \uparrow $             & ${\text{mAP}_{\texttt{known}}} \uparrow $           \\ 
\midrule
BEVFormer (closed\text{-}set)   & 0             & 0             & 43.0                          & 0                & 0                & 43.1          \\
BEVFormer (Supervised)          & 52.5          & 20.5          & 44.6                          & 52.3             & 42.6             & 43.1          \\ 
\midrule
DETR3D$+$OW\text{-}DETR      & 12.7          & 0             & 38.9                          & 0                & 0                & 31.8          \\
BEVFormer$+$OW\text{-}DETR     & 0.3           & 0             & 44.1                          & 1.0                & 0                & 37.1          \\
BEVFormer$+$CA-3D      & 16.7          & 0.7           & 44.3                          & 25.9             & 1.4              & 42.5          \\
\midrule
\textbf{BEVFormer}\textbf{+OS-Det3D (ours)}    & \textbf{23.2} & \textbf{0.7}  & \textbf{45.1}  & \textbf{31.8}    & \textbf{4.2}     & \textbf{43.4} \\ 
\bottomrule[1.2pt]
\end{tabular}
}
}
\label{tab:nusc-os}
\end{table*}

\subsection{Implementation}
\subsubsection{Datasets} Our method was evaluated on two representative benchmarks for autonomous driving: nuScenes \cite{nuscenes2019} and KITTI \cite{geiger2012we}. Notably, in the real world, known and unknown instances often appear simultaneously, and the outdoor approach \cite{cen2021open, cen2022open} for handling known and unknown instances differs in terms of whether their annotations are retained. Like in the MLUC \cite{cen2021open}, in KITTI, 3 common classes (\textit{car}, \textit{pedestrian}, \textit{cyclist}) are classified as known classes, whereas (\textit{van}, \textit{truck}) are used as the unknown classes. For the nuScenes dataset, we define 2 different dataset splits. In \textbf{nuScenes Split 1}, we classify 3 common classes (\textit{car}, \textit{pedestrian}, \textit{bicycle}), while the remaining 8 classes (\textit{barriers}, \textit{construction vehicles}, \textit{truck}, \textit{bus}, \textit{trailer}, \textit{motorcycle}, \textit{traffic cone}, \textit{debris}) are treated as unknown categories.

Typically, the geometry-based region proposal method performs better when there is a richer diversity of object categories. To validate this hypothesis, we designed the \textbf{nuScenes Split 2} dataset setup and assessed the robustness of our method under different configurations within the nuScenes dataset. We also reference the \textbf{KITTI split}, which involves including unknown categories that are geometrically similar to the known categories in the unknown split while maintaining a similar ratio of known to unknown objects in the \textbf{nuScenes Split 2} validation dataset for testing as that in the \textbf{KITTI Split}. In \textbf{nuScenes Split 2}, we add (\textit{barriers}, \textit{construction vehicles}) to the known classes, whereas the remaining 6 classes (\textit{truck}, \textit{bus}, \textit{trailer}, \textit{motorcycle}, \textit{traffic cone}, \textit{debris}) are designated as unknown categories. The detailed split settings can be found in Table \ref{tab:task_split}.

\subsubsection{Evaluation Metrics} For the KITTI dataset, we adopted the evaluation metrics of the MLUC \cite{cen2021open} by using the mean average precision $(\text{mAP}_{\texttt{known}})$, unknown recall $(\text{Recall}_{\texttt{unk}})$ and unknown average precision $(\text{AP}_{\texttt{unk}})$. Both the evaluation difficulty level, which was set to moderate, and the intersection over union (IoU) thresholds (0.7 for cars and trucks, 0.5 for pedestrians and cyclists, and 0.1 for unknown objects such as vans and trucks) are consistent with the MLUC setup. Similar to the nuScenes dataset, we utilized $\text{mAP}_{\texttt{known}}$ to assess the performance for known objects while reporting the average recall $(\text{AR}_{\texttt{unk}})$ and $\text{AP}_{\texttt{unk}}$ for the unknown objects. These metrics were evaluated on the basis of the standard nuScenes benchmark \cite{nuscenes2019}, and the distance thresholds for both known and unknown categories are $(0.5m, 1m, 2m, 4m)$, following the standard nuScenes benchmark parameter settings.

\subsubsection{Model Details}

Most of the basic hyperparameter settings can be found in the referenced works related to BEVFormer and Object DGCNN~\cite{obj-dgcnn, li2022bevformer}; here, we focus specifically on describing the most critical hyperparameters used in our framework. The parameters $\tau_{1}$ and $\tau_{2}$ in the Gaussian kernel functions $\phi_1$ and $\phi_2$ were set to $0.5$ and $0.05$, respectively. In the first stage, ODN3D was trained for 20 epochs in a class-agnostic manner using only instances of known classes, with a batch size of 4 per GPU and an initial learning rate of $1\times10^{-4}$. BEVFormer served as our camera baseline and was trained under a standard closed-set setting for 18 epochs with a batch size of 1 per GPU and an initial learning rate of $2\times10^{-4}$. In the second stage, we reloaded the pretrained weights from BEVFormer, which were obtained from the closed-set setting (excluding the classification branch), and further trained it under the OS-Det3D framework for an additional 6 epochs, maintaining a batch size of 1 per GPU and an initial learning rate of $2\times10^{-4}$. For both stages, the AdamW optimizer was used. The hyperparameters $k_{o}$ and $k_{u}$ were set to 30 and 10, respectively. All of the models across both stages were trained on 8 Tesla V100 GPUs.

\subsection{Open-set 3D Object Detection Results}
To demonstrate the effectiveness of our method, we conducted experiments on the nuScenes \cite{nuscenes2019} and KITTI \cite{geiger2012we} datasets. Following the category splitting settings described above, we divided both the nuScenes training and validation sets into two data partitions: nuScenes Split 1 and nuScenes Split 2.

\begin{table}[t]
    \label{tab:kitti}
    \centering
    \caption{Comparison of unknown discovery task on the KITTI test set.  \\ $\dagger$ indicates the results from their respective papers.}
    {\footnotesize\setlength{\tabcolsep}{.8mm}{\begin{tabularx}{0.8\columnwidth}{@{}l|ccc@{}}
        \toprule[1.2pt]
        \textbf{Methods} & $\text{Recall}_{\texttt{unk}} \uparrow $  & $\text{AP}_{\texttt{unk}} \uparrow $    & $\text{mAP}_{\texttt{known}} \uparrow $ \\
        \midrule
        SECOND (Closed-set)                   & 0               & 0             & 67.4  \\
        SECOND (Supervised)                             & 94.3            & 76.4          & 72.6 \\ 
        \midrule
        MC-Dropout $\dagger$  & -               & 2.6           & 64.1 \\
        OSIS $\dagger$     & 31.0            & 1.1           & 65.9\\
        MLUC   $\dagger$            & 50.0            & 9.7           & 66.8  \\ 
        \midrule
        ODN3D                                    & \textbf{74.4}  & 3.0           & \textbf{74.5}\\
        ODN3D + GT Filtering                     & \textbf{74.4}   & \textbf{33.2} & \textbf{74.5}\\ 
        \bottomrule[1.2pt]
        \end{tabularx}
        }
        }
\end{table}

\subsubsection{Results on the nuScenes Dataset}
Table \ref{tab:nusc-os} compares two pseudo-label generation approaches with our proposed OS-Det3D method. +OW-DETR refers to a feature-based pseudo-labeling method adapted from OW-DETR \cite{gupta2022ow}, and CA-3D serves as a baseline class-agnostic 3D detector based on Object DGCNN \cite{obj-dgcnn}, which uses binary classification to generate object proposals. BEVFormer (closed-set) indicates that BEVFormer was trained with only known category labels, serving as a baseline for the $\text{mAP}_{\texttt{known}}$ values. BEVFormer (Supervised) includes both known and unknown category labels, providing an upper bound for unknown detection performance in terms of the $\text{AR}_{\texttt{unk}}$ and $\text{AP}_{\texttt{unk}}$ values.
DETR3D+OW-DETR indicates that the 2D plane features were extracted from the image backbone, whereas BEVFormer+OW-DETR indicates that the BEV features were extracted from the transformer encoder. These different feature extraction mechanisms influence the quality of the pseudo-labels generated during training. Compared with the CA-3D method, OS-Det3D improves the unknown detection performance; with a $\text{AR}_{\texttt{unk}}$ increase of 6.5\% on nuScenes Split 1 and a $\text{AR}_{\texttt{unk}}$ increase of 5.2\% on the nuScenes Split 2 test set, it also performs well in terms of detecting known objects.

\begin{figure*}[t]
\centering
  \includegraphics[width=1\textwidth]{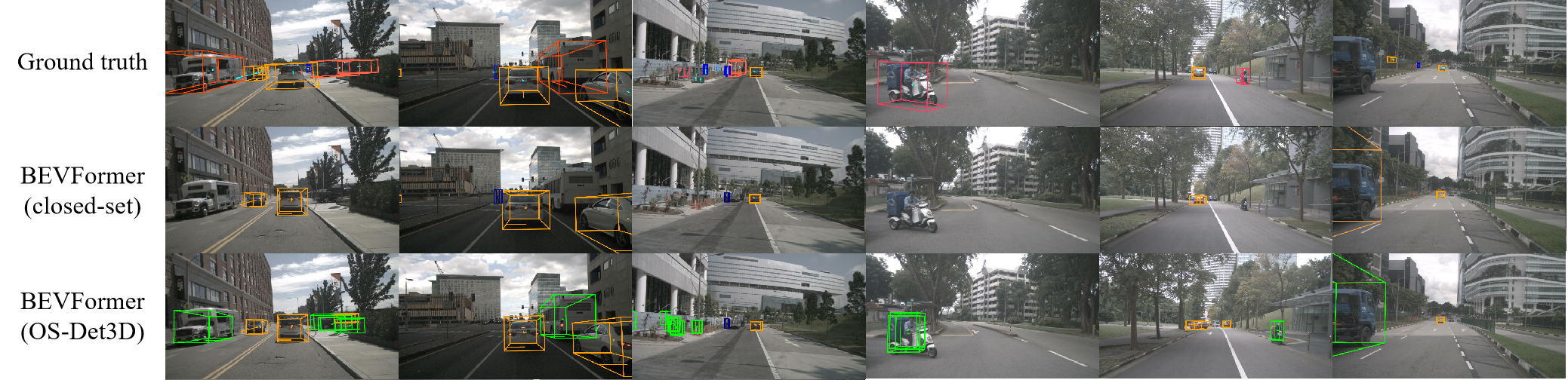}
  \caption{ \textbf{Visualization results on the nuScenes Split 2 test set}. The qualitative results showcase different outcomes. Row 1 corresponds to the ground truth with all classes: car, pedestrian, bicycle, barrier, construction vehicle, truck, bus, trailer, motorcycle, traffic cone, and debris (trash bins, etc.). Row 2 presents the results from BEVFormer (closed-set), which focuses only on detecting known classes (car, pedestrian, bicycle, barrier, and construction vehicle), and Row 3 presents the results obtained by our BEVFormer+OS-Det3D, which is able to identify unknown instances as green (truck, bus, trailer, motorcycle, traffic cone, and debris). These images have been zoomed in for a better view.
  }
  \label{fig:qr}
\end{figure*}

\subsubsection{Results on the KITTI Dataset}
Since the open-set 3D object detection benchmark in KITTI involves only LiDAR 3D detectors, we present the ODN3D results to provide a fair comparison in Table \ref{tab:kitti} by following the training and evaluation protocol of the MLUC \cite{cen2021open}. ODN3D+GT Filtering evaluates all 3D region proposals as unknown objects and filters out the prediction overlap with the ground truth of known categories before evaluation. Our results show that ODN3D improves the $\text{Recall}_{\texttt{unk}}$ values when compared to all of the baselines, demonstrating superior unknown instance discovery. Additionally, ODN3D outperforms the MLUC in terms of the $\text{AP}_{\texttt{unk}}$ by 23.5\%, highlighting the effectiveness of our framework in pseudo-labeling tasks.

\subsection{Ablation Study}

\begin{table}[t]
    \centering
    {\caption[Valori medi]{Ablation of OS-Det3D on the nuScenes Split 2 test set. JS refers to the joint selection module, CA-3D serves as a baseline class-agnostic 3D detector based on Object DGCNN, GF refers to GT filtering, and SW refers to soft weighting.}
    {\footnotesize\setlength{\tabcolsep}{.8mm}{\begin{tabularx}{0.8\columnwidth}{@{}c|ccc|ccc@{}}
    \toprule[1.2pt]
    \textbf{ID}   & Stage1 & Stage2 & SW & $\text{AR}_{\texttt{unk}} \uparrow $  & $\text{AP}_{\texttt{unk}} \uparrow $    & $\text{mAP}_{\texttt{known}} \uparrow $ \\
    \midrule
    1  & -  & - & -  & 0 & 0 & 43.1 \\
    2  & CA-3D & GF & -  & 25.9 & 1.4 & 42.5 \\
    3  & ODN3D & GF & -  & 26.6 & 3.3 & 42.0 \\
    4  & ODN3D & JS & - & 31.7 & 2.1 & 42.5 \\
    5  & ODN3D & GF & \checkmark & 26.9 & 1.8 & 41.2 \\
    6  & ODN3D & JS & \checkmark  & \textbf{31.8} & \textbf{4.2} &  \textbf{43.4} \\
    \bottomrule
      \end{tabularx}}}
      \label{tab:OS-Det3D-ab}}
\end{table}

\subsubsection{Ablation of the OS-Det3D Components}
We conducted ablation experiments on OS-Det3D to evaluate the impact of different components on nuScenes Split 2, as shown in Table \ref{tab:OS-Det3D-ab}. The standard BEVFormer model serves as the baseline, demonstrating the basic detection performance for known classes. Initially, incorporating only ODN3D predictions as pseudo-labels for unknown objects enhances open-set detection but reduces known class detection performance by 7.9\% $\text{mAP}_{\texttt{known}}$. By introducing the joint selection module, we enhance the detection performance with a 10.1\% improvement in the $\text{Recall}_{\texttt{unk}}$ and a 7.3\% increase in the $\text{mAP}_{\texttt{known}}$. Soft weighting (SW) specifically refers to the process in which $s_{obj}$ is used as a weight for the pseudo ground truth when calculating the classification loss, which further enhances the performance for both known and unknown categories. The best results are achieved by combining all of the components.

\subsubsection{Ablation of ODN3D}
To study the contribution of each component in ODN3D, we present the ablation study results in Table \ref{tab:LiDAR-rpn-ab}. Object DGCNN trained with binary classification serves as our baseline. Replacing Hungarian matching with GeoHungarian matching in the baseline and updating the model by calculating the objectness and regression loss significantly reduces the detection performance, and when including all of our designs, ODN3D achieves the highest $\text{AR}_{\texttt{unk}}$. To further validate ODN3D's effectiveness in generic 3D object detection, we trained it on all nuScenes categories as a single class, and ODN3D exhibits improvements in terms of both the AR and AP over the baseline for detecting general 3D objects.

\begin{figure*}[t]
\centering
  \includegraphics[height=6.5cm]{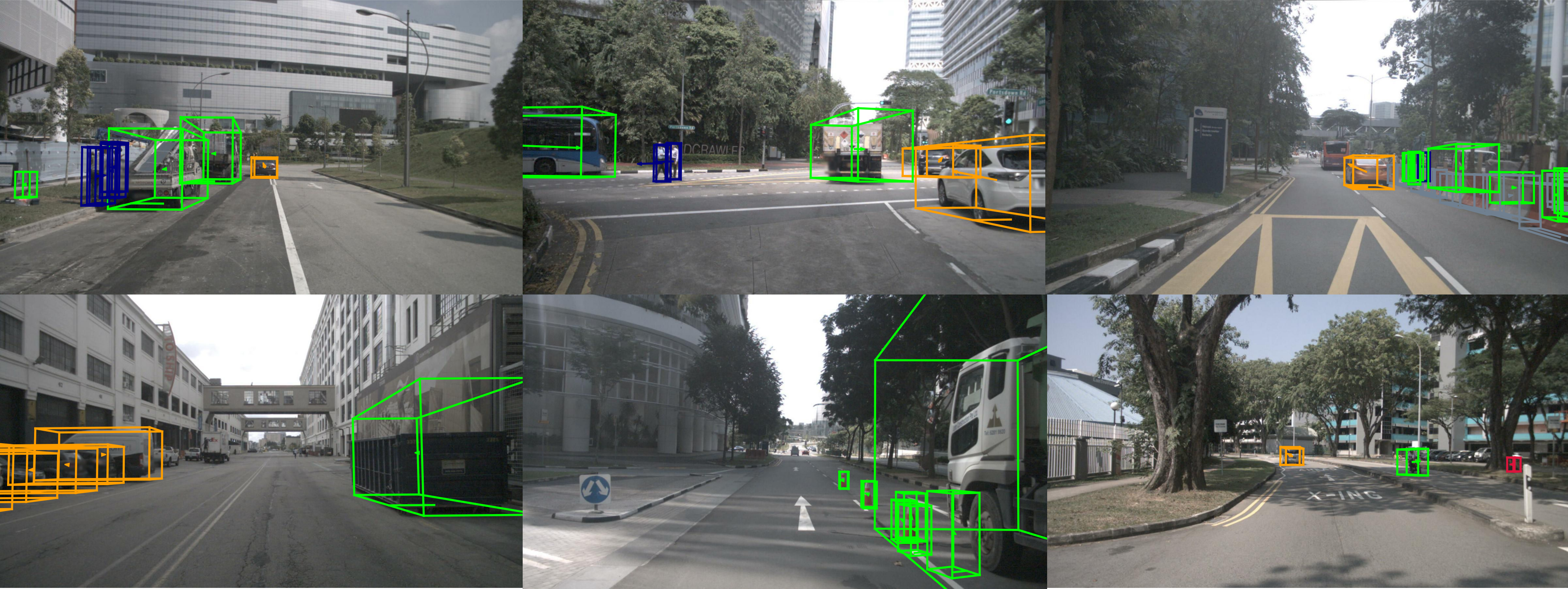}
  \caption{ \textbf{Additional visualization results using BEVFormer (OS-Det3D) on nuScence Split 1 test set}. Known classes: \textit{car (yellow), pedestrian (blue), bicycle (red)}; Unknown classes (green): bus, truck, traffic cone, trailer, motorcycle, debris. These images have been zoomed in for a better view.
  }
  \label{fig:m-qr}
\end{figure*}

\begin{table}[t]
    \centering
    {\caption[Valori medi]{Ablation of ODN3D on the test set. nuScenes (all) indicates that all categories are treated as a single category during training. H, GH, and OBJ refer to Hungarian 3D matching, GeoHungarian matching and the 3D objectness scores.}
    {\footnotesize\setlength{\tabcolsep}{.8mm}{\begin{tabularx}{0.7\columnwidth}{@{}c|ccc|cc|cc@{}}
        \toprule[1.2pt]
        \multirow{2}{*}{\textbf{ID}} & \multicolumn{3}{c|}{}  & \multicolumn{2}{c|}{\textbf{nuScenes Split 2}}  & \multicolumn{2}{c}{\textbf{nuScenes (all)}}  \\
         & H & GH & OBJ & $\text{AR}_{\texttt{unk}} \uparrow $ & $\text{AP} \uparrow $ & $\text{AR} \uparrow $ & $\text{AP} \uparrow $ \\ 
        \midrule
        1  & \checkmark     & -          & -          & 50.1           & \textbf{62.2} & 85.0 & 67.0\\
        2  & -              & \checkmark & \checkmark & 50.9           & 4.2           &  -   &  - \\
        3  & \checkmark     & -          & \checkmark & 49.8           & 58.7          &  -   &  - \\
        4  & \checkmark     & \checkmark & \checkmark & \textbf{56.4}  & 58.6          & \textbf{85.9} & \textbf{70.2} \\ 
        \bottomrule[1.2pt]
      \end{tabularx}}}
      \label{tab:LiDAR-rpn-ab}}
\end{table}

\begin{table}[t]
    \centering
    {\caption[Valori medi]{Comparison of score designs on the nuScenes Split 2 test set. * indicates that we adopt the OLN objectness score for the 3D, and PCA represent principal component analysis.}
    {\footnotesize\setlength{\tabcolsep}{1mm}{\begin{tabularx}{0.9\columnwidth}{@{}l|cc||l|ccc@{}}
    \toprule[1.2pt]
    \multicolumn{3}{c||}{3D Objectness Score} & \multicolumn{4}{c}{Joint Selection Score} \\
    \midrule
    \text{Methods} & $\text{AR}_{\texttt{unk}} $ & $\text{AP} $ & \text{Scores} & $\text{AR}_{\texttt{unk}}  $  & $\text{AP}_{\texttt{unk}}  $    & $\text{mAP}_{\texttt{known}} $ \\
    \midrule
    IOU3D  &  48.2   &  52.0 & $s^{\prime}_{obj}$  & 29.8 & 3.2 & 40.9 \\
    RIOU3D &  50.0   &  52.0 & $s_{fea}$ & 29.7 & 2.7 & 43.2\\
    OLN*      &  50.8   &  47.0 & $s_{jos}$ (PCA)  & 31.0 & \textbf{5.2} &  43.2 \\
    Ours &  \textbf{56.4}   &  \textbf{58.6} & $s_{jos}$ (mean)  & \textbf{31.8} & 4.2 &  \textbf{43.4} \\
    \bottomrule[1.2pt]
      \end{tabularx}}}
      \label{tab:object-score}}
\end{table}

\begin{table}[t]
    \centering
    {\caption[Valori medi]{Sensitivity Analysis of $k_{u}$ on the nuScenes Split 2 test set.}
    {\footnotesize\setlength{\tabcolsep}{1mm}{\begin{tabularx}{0.55\columnwidth}{@{}c|ccc@{}}
        \toprule[1.2pt]
        $k_{u}$ & $\text{mAP}_{\texttt{known}}\uparrow$ & $\text{AP}_{\texttt{unk}}\uparrow$ & $\text{AR}_{\texttt{unk}}\uparrow$ \\
        \midrule
        3    & 43.1     & 1.1     & 24.3       \\
        5    & \textbf{43.6}     & 2.4     & 29.8       \\
        10   & 43.4     & \textbf{4.2}     & \textbf{31.8}       \\
        20   & 43.1     & 3.5     & 30.9       \\
        30   & 42.0     & 3.3     & 30.2       \\ 
        \bottomrule[1.2pt]
      \end{tabularx}}}
      \label{tab:sensitivity-table}}
\end{table}

\subsubsection{Discussion of Score Designs}
Table \ref{tab:object-score} presents the results of the 3D Objectness Score and Joint Selection Score Designs on the nuScenes Split 2 test set.

In terms of the objectness score design, we experimented with different approaches. In addition to IOU3D \cite{shi2019pointrcnn} and RIOU3D \cite{zheng2020rotation}, we adopted the OLN objectness score for the 3D by computing the geometric means of the centerness and IOU3D. The results demonstrate that our design for the 3D objectness score, which effectively integrates the rotation angle information and calculates the centerness and scale separately, leads to superior performance.

In terms of the joint selection score design, we observe that if the pseudo-GT contains many object boxes that are very similar to known semantics, it will undermine the detector's performance. Furthermore, if the pseudo-GT contains inaccurate 3D information, the model's ability to learn and detect unknown objects will be compromised. Experiments with the joint selection score design reveal that combining cross-modality information with pseudo-label selection significantly outperforms relying solely on scores from a single modality in our approach.

\subsubsection{Sensitivity Analysis of $k_{u}$ from the Joint Selection Module}
As shown in Table \ref{tab:sensitivity-table}, we analyzed the model's detection performance on nuScenes Split 2 under different settings of the hyperparameter $k_{u}$ while keeping $k_{o}=30$. A larger value of $k_{u}$ negatively impacts the $\text{mAP}_{known}$ values, whereas a smaller $k_{u}$ value results in insufficient unknown samples for effective training. Based on the validation results, we set $k_{u}$ to 10 to achieve a balanced performance between known and unknown categories.

\subsection{Visualization}

\begin{figure*}[ht]
  \centering
  \includegraphics[height=6.5cm]{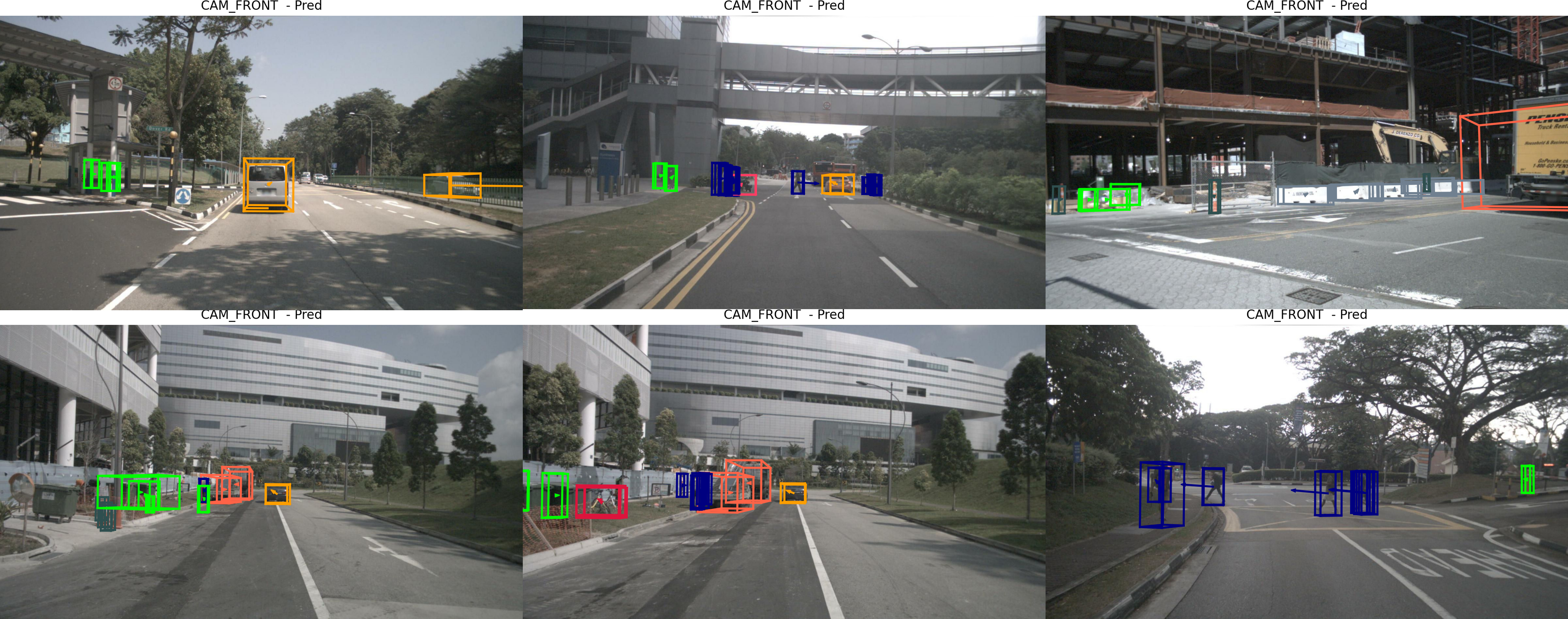}
  \caption{\textbf{Discovering debris with BEVFormer (OS-Det3D) on the nuScenes(all) test set}. Known classes: \textit{car (yellow), pedestrian (blue), bicycle (red), bus, truck (orange), traffic cone (dark green), trailer, motorcycle (pink)}; Unknown classes (green): \textit{debris}. These images have been zoomed in for a better view.
}
  \label{fig:debris}
\end{figure*}

In Fig.~\ref{fig:qr}, we present the qualitative results of OS-Det3D on the nuScenes Split 2 test set. When trained under the conventional closed-set paradigm, BEVFormer (closed-set) fails to detect unknown objects on the road. In contrast, BEVFormer (OS-Det3D), which was trained with our proposed framework, accurately localizes previously unseen objects such as trucks, buses, and motorcycles that are absent from the training set. Additionally, we present the results of OS-Det3D on the nuScenes Split 1 test set in Fig.~\ref{fig:m-qr}, further demonstrating its effectiveness across different data partitions and more quantitative results. We also conducted a visual analysis on challenging corner-case scenarios. As illustrated in Fig.~\ref{fig:debris}, our model exhibits strong detection responses to unusual or miscellaneous objects, such as trash bins along the roadside, highlighting its robustness in complex open-world environments.

\subsection{Discussion \& Further Analysis}
\begin{table}[t]
    \centering
    {\caption[Valori medi]{Extended training with a greater number of epochs on nuScenes Split 2.}
    {\footnotesize\setlength{\tabcolsep}{1mm}{\begin{tabularx}{0.8\columnwidth}{@{}l|ccc@{}}
        \toprule[1.2pt]
         Method & $\text{mAP}_{\texttt{known}}\uparrow$ & $\text{AP}_{\texttt{unk}}\uparrow$ & $\text{AR}_{\texttt{unk}}\uparrow$ \\
        \midrule
        BEVFormer (24 epochs)  & 43.1  & -   & -      \\
        OS-Det3D (24 epochs)   & 43.4  & 4.2 & 31.8       \\
        \midrule
        BEVFormer (30 epochs)  & 43.0  & -   & -       \\
        OS-Det3D (30 epochs)   & \textbf{44.6}  & 5.1 & 31.9       \\
        \midrule
        BEVFormer (36 epochs)  & 43.2  & -   & -  \\
        OS-Det3D (36 epochs)   & 43.9  & \textbf{5.3} & \textbf{32.7}       \\ 
        \bottomrule[1.2pt]
      \end{tabularx}}}
      \label{tab:highepoch}}
\end{table}

\begin{figure*}[!t]
  \centering
    \subfloat[]{\includegraphics[width=3in]{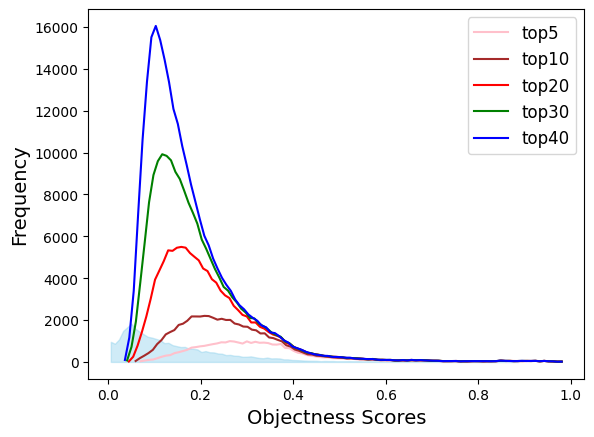}
    \label{fig:short-a}}
    \hspace{15mm}
    \subfloat[]{\includegraphics[width=3in]{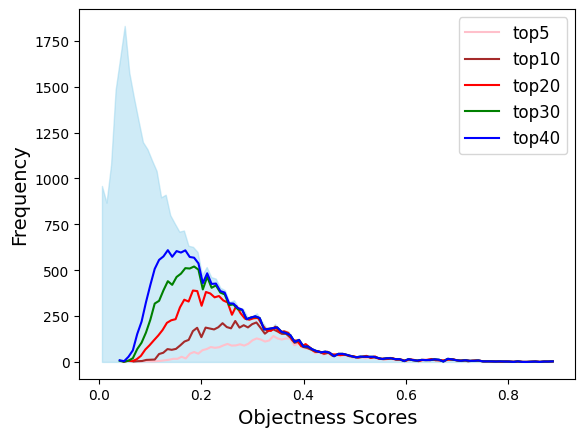}
    \label{fig:short-b}}
  \caption{Analysis of $k_{o}$. The curve in Fig. (a) represents the number of proposals provided by ODN3D under different $k_{o}$ selections, whereas the curve in Fig. (b) represents the unknown ground truth that can be matched under different $k_{o}$ selections. The light blue shaded area represents the number of proposals from ODN3D that match instances of the unknown ground truth in the nuScenes Split 2 test set as well as the distribution of the scores. In Fig. a and b, the sizes of the shaded areas are the same. These images have been zoomed in for a better view.
}
  \label{fig:gt-filtering}
\end{figure*}

\subsubsection{Extended Training Over More Epochs}

In the standard BEVFormer training setup, 24 epochs are typically used. In our approach, OS-Det3D follows a two-phase training strategy: closed-set training for the first 18 epochs followed by open-set training for the remaining 6 epochs, maintaining a total of 24 epochs to allow for a fair comparison. We then extended the training beyond 24 epochs to investigate its effects, as shown in Table \ref{tab:highepoch}. We observed that while the loss continued to gradually decrease after 24 epochs, it had not fully converged. Only after an additional 12 epochs (reaching the 36th epoch) did the loss show signs of convergence and the model's performance stabilize. Notably, while the standard BEVFormer model converged by the 24th epoch and showed no further gains with additional training, our approach demonstrated superior performance with extended training.

\begin{table}[t]
\centering
\caption{ Proportion of matched candidates in the total candidates when the ground truth for the unknowns is visible.}
\begin{tabular}{c|ccccc}
\hline
$k_{o}$      & 5    & 10   & 20  & 30  & 40  \\ \hline
match/total(\%)         & 12.4 & 10.4 & 8.2 & 7.0 & 6.1 \\ \hline
\end{tabular}
\label{tab:topk}
\end{table}

\subsubsection{Sensitivity Analysis of $k_{o}$ from the GT Filtering Process}

Since the ground truth for the known instances is available, we can perform GT filtering directly. This process was conducted frame by frame, where 3D object region proposals that overlapped with known class objects were removed. Fig. \ref{fig:gt-filtering} shows the analysis of the different $k_{o}$ selected proposals sorted by $s_{obj}$. The $\text{top-}k_{o}$ proposals after GT filtering indicate the number of 3D region candidates that obtain a potential unknown object. From each frame, we extracted the top-$k_{o}$ proposals as 3D object region candidates, which were then input into the joint selection module. As shown in Fig. \ref{fig:short-a}, each curve depicts the distribution of the 3D object region candidates across different score values, with the area under the curve representing the total number of extracted 3D region candidates. Correspondingly, in Fig. \ref{fig:short-b}, each curve displays the number of 3D object region candidates matching the ground truth across different score distributions. Thus, the area under each curve represents the total number of 3D object region candidates matching the ground truth. Since the selection of $k_{o}$ is related to the unknown ground truth, we did not choose a value with a high proportion, as shown in Table \ref{tab:topk}. Instead, we selected a lower value to demonstrate the effectiveness of our method.

\section{Conclusions}
This work presents OS-Det3D, a novel framework that targets the critical challenge of open-set 3D object detection when using cameras in autonomous driving scenarios. Traditional camera-based 3D detectors are constrained by the closed-set assumption, limiting their ability to handle unknown objects in real-world environments. OS-Det3D addresses this limitation by implementing a two-stage design that integrates a 3D object discovery network (ODN3D) for class-agnostic proposal generation and a joint selection module for filtering and identifying unknown objects. This approach enables camera-only 3D detectors to not only detect and localize known categories but also recognize previously unseen targets. Extensive experiments across multiple benchmarks demonstrate the effectiveness and robustness of OS-Det3D in terms of improving the detection performance for both known and unknown objects.

\textbf{Limitations and Future Work.} While OS-Det3D exhibits a strong performance in terms of detecting both known and unknown objects across diverse datasets, several limitations remain. First, the current framework relies on a 3D objectness scoring mechanism that assumes a certain degree of scale similarity between known and unknown objects. This assumption may affect generalization when unknown targets exhibit significant size variations. Future work will explore scale-invariant geometric representations and adaptive objectness scoring strategies to address this limitation.
Second, although OS-Det3D incorporates LiDAR data during training via the ODN3D module, the current inference pipeline is camera-only. In future work, we plan to investigate full multimodal inference, e.g., by fusing LiDAR and camera BEV features, to further enhance the robustness of open-set detection under challenging conditions.

\bibliographystyle{IEEEtran}
\bibliography{IEEEfull}

\vspace{-12 mm} 

\begin{IEEEbiography}[{\includegraphics[width=1.in,height=1.25in,clip,keepaspectratio]{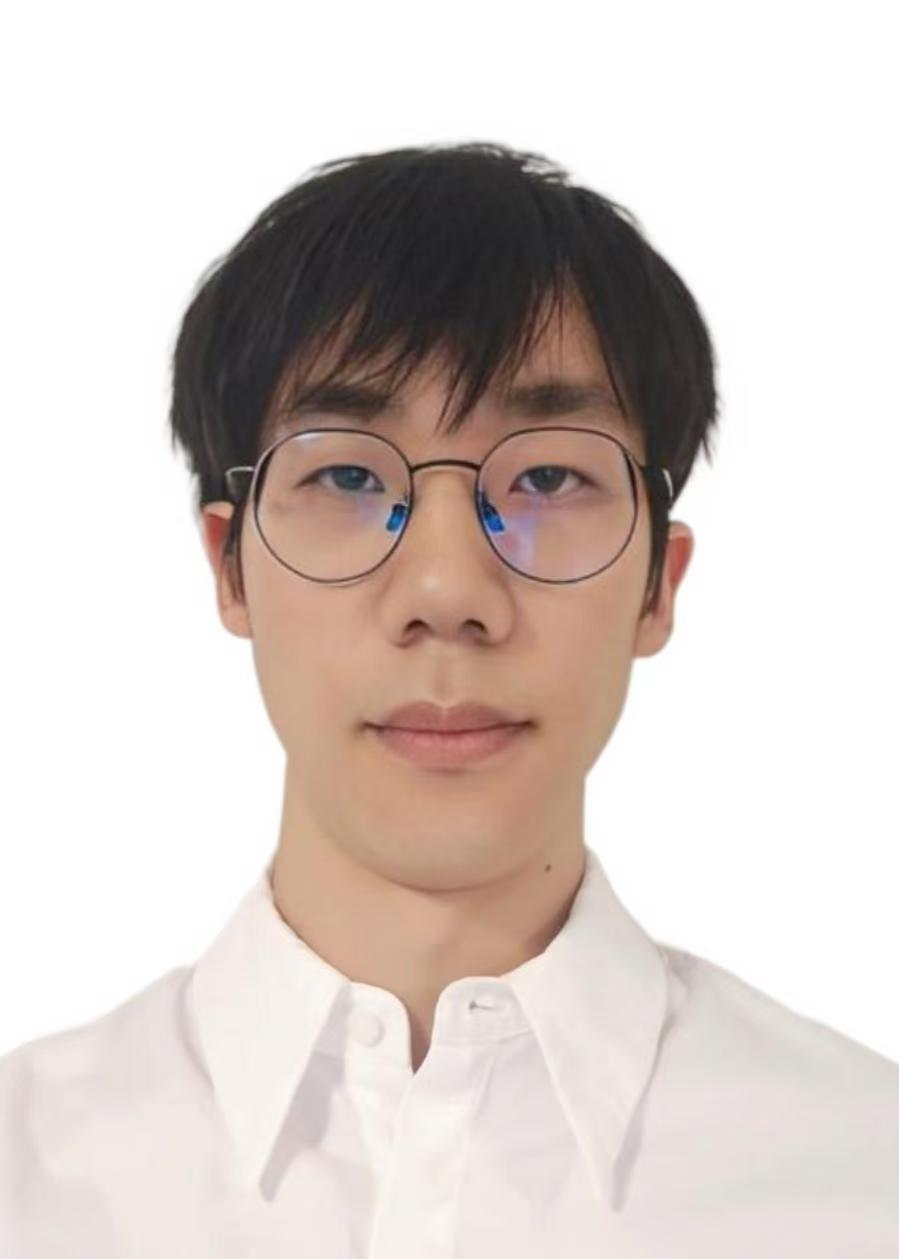}}]{Zhuolin He} received his  B.S degree from Dalian University of Technology in 2020. He is currently pursuing the Ph.D. degree with the School of Computer Science, Fudan University, Shanghai, China. His current research interests include deep learning, computer vision and 3D object detection.
\end{IEEEbiography}

\vspace{-12 mm} 

\begin{IEEEbiography}
[{\includegraphics[width=1.in,height=1.25in,clip,keepaspectratio]{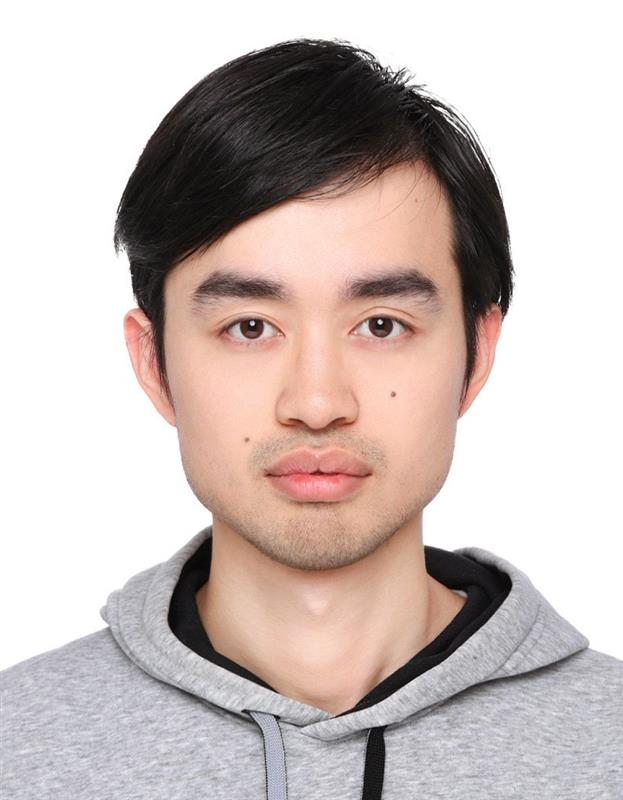}}]
{Xinrun Li} completed his bacholar degree at Tongji University in 2020. After graduating, he joined Bosch Corporate Research. His research interests include 3D perception and sensor fusion, primarily focusing on stastic environment perception in autonoumous driving.
\end{IEEEbiography}

\vspace{-12 mm} 

\begin{IEEEbiography}[{\includegraphics[width=1.in,height=1.25in,clip,keepaspectratio]{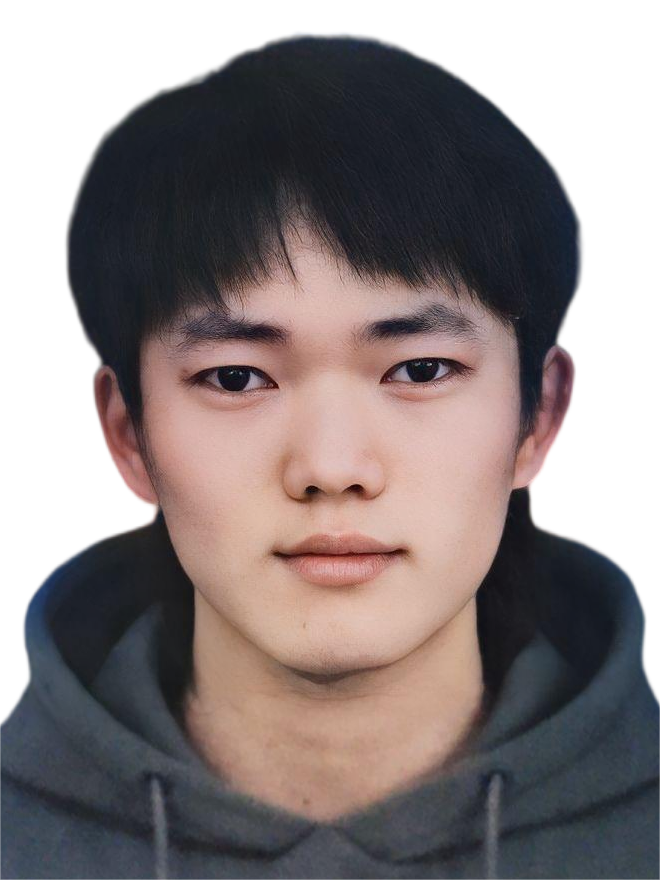}}]{Jiacheng Tang} received the B.Eng. degree from Xidian University in 2021. He is currently pursuing a Ph.D. degree at the Institute of Science and Technology for Brain-Inspired Intelligence (ISTBI), Fudan University, Shanghai, China. His current research interests focus on deep learning and its applications in computer vision and autonomous driving.
\end{IEEEbiography}

\vspace{-12 mm} 

\begin{IEEEbiography}[{\includegraphics[width=1.in,height=1.25in,clip,keepaspectratio]{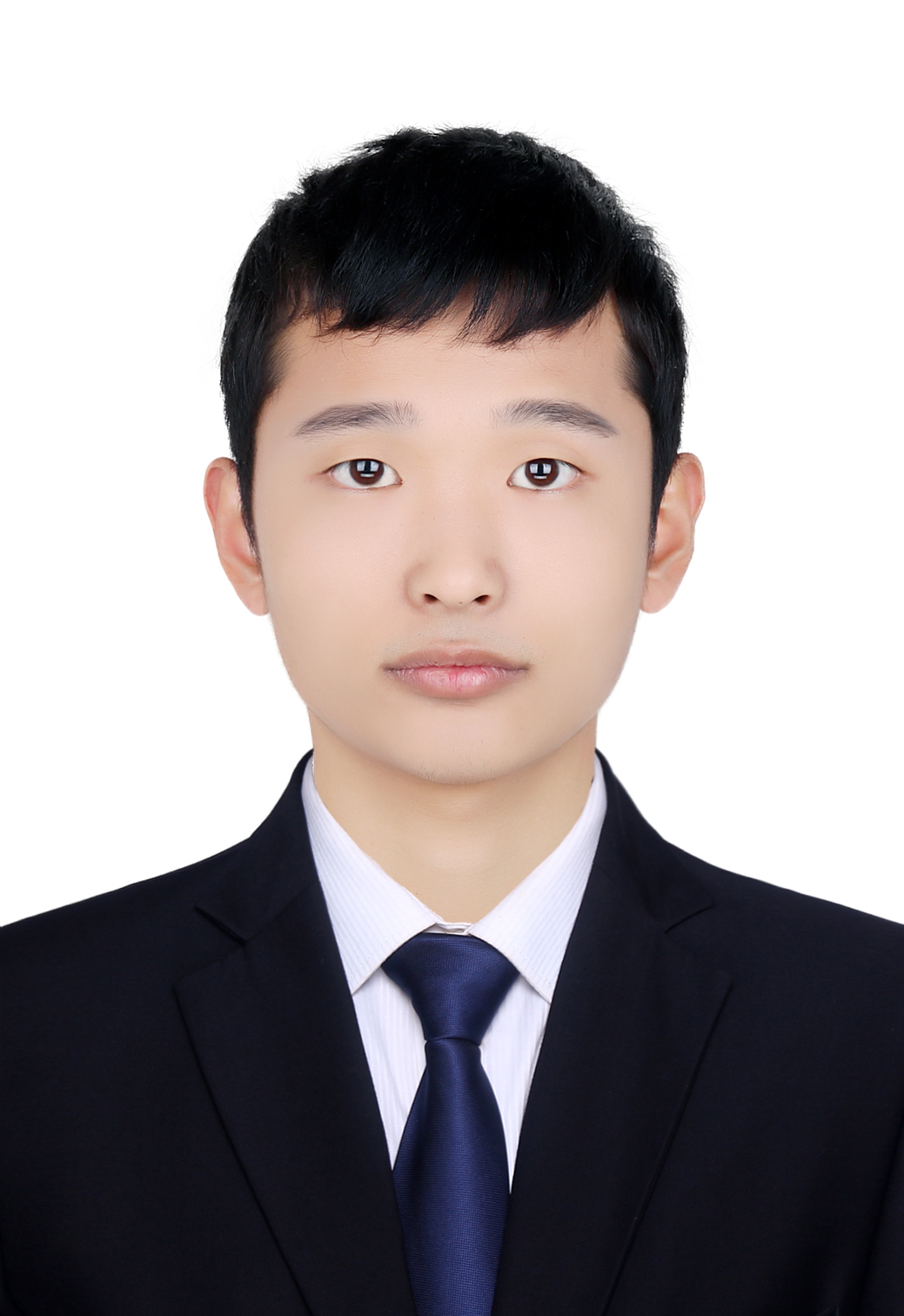}}]{Shoumeng Qiu} received the Master's degree from Shanghai Institute of Microsystem and Information Technology, Chinese Academy of Sciences, and University of Chinese Academy of Sciences, China, in 2021. He is currently a Ph.D. student at Fudan University. His current research interests include 3D point cloud understanding, semantic segmentation, machine learning and artificial intelligence.
\end{IEEEbiography}

\vspace{-12 mm}

\begin{IEEEbiography}[{\includegraphics[width=1.in,height=1.25in,clip,keepaspectratio]{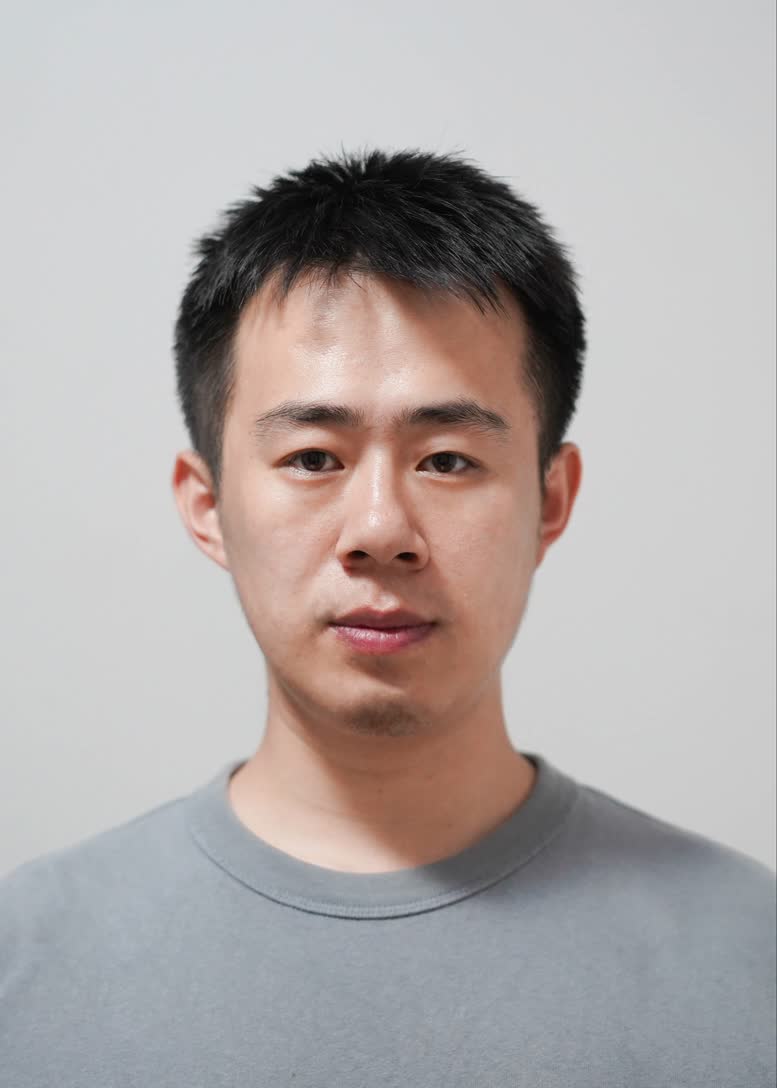}}]{Wenfu Wang} completed his doctoral thesis at Zhejiang University in 2020. After graduating, he joined  Autonomous Driving (AD) company to engage in research and development work related to 3D object detection. In 2022, he joined Bosch Corporate Research. His research interests include 3D computer vision, primarily focusing on 3D object detection and 3D representation learning in AD. Recently, he has been interested in foundational models for 3D vision and 3D vision in MLLM (Multimodal Large Language Models).
\end{IEEEbiography}

\vspace{-12 mm} 

\begin{IEEEbiography}[{\includegraphics[width=1.in,height=1.25in,clip,keepaspectratio]{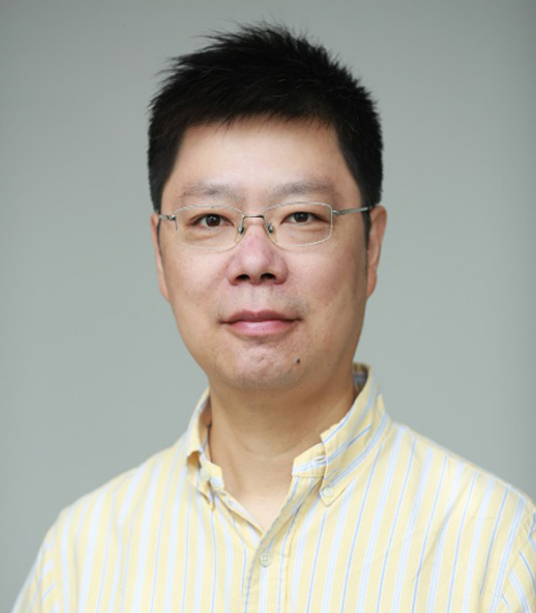}}]{Xiangyang Xue} received the BS, MS, and PhD degrees in communication engineering from Xidian University, Xi’an, China, in 1989, 1992, and 1995, respectively. He is currently a professor of computer science with Fudan University, Shanghai, China. His research interests include multimedia information processing and machine learning.
\end{IEEEbiography}

\vspace{-12 mm} 

\begin{IEEEbiography}[{\includegraphics[width=1.in,height=1.25in,clip,keepaspectratio]{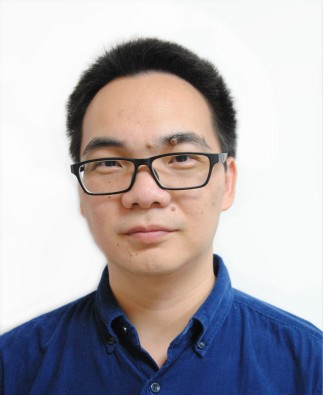}}]{Jian Pu} received a Ph.D. degree from Fudan University, Shanghai, China, in 2014. Currently, he is a Young Principal Investigator at the Institute of Science and Technology for Brain-Inspired Intelligence (ISTBI), Fudan University. He was an associate professor at the School of Computer Science and Software Engineering, East China Normal University from 2016 to 2019, and a postdoctoral researcher of the Institute of Neuroscience, Chinese Academy of Sciences in China from 2014 to 2016. His current research interest is to develop machine learning and computer vision methods for autonomous driving.
\end{IEEEbiography}

\end{document}